\documentclass[sigconf]{acmart}

\usepackage{booktabs} 
\usepackage{times}
\usepackage{xcolor}
\usepackage{soul}
\usepackage{graphics}
\usepackage{graphicx}
\usepackage{subfig}
\usepackage{amsfonts}
\usepackage{algorithm}
\usepackage[export]{adjustbox}
\usepackage{array}
\usepackage[noend]{algpseudocode}
\usepackage[utf8]{inputenc}

\editor{Anoop M. Namboodiri}
\editor{Vineeth Balasubramanian}
\editor{Amit Roy-Chowdhury}
\editor{Guido Gerig}

\newcommand*{\origrightarrow}{}

\renewcommand*{\textrightarrow}{\fontfamily{cmr}\selectfont\origrightarrow}

\begin{document}
\title{Learning to Prevent Monocular SLAM Failure using Reinforcement Learning}
\subtitle{}



\author{Vignesh Prasad}
\affiliation{%
  \institution{Embedded Systems \& Robotics\\TCS Research \& Innovation}
  \city{Kolkata, India}
}
\email{vignesh.prasad@tcs.com}

\author{Karmesh Yadav}
\affiliation{%
  \institution{Robotics Institute \\Carnegie Mellon University}
  \city{Pittsburgh, USA}
}
\email{ykarmesh@gmail.com}

\author{Rohitashva Singh Saurabh}
\affiliation{%
  \institution{Dept. of Robotics Engineering \\John Hopkins University}
  \city{Baltimore, USA}
}
\email{saurabhiitg11@gmail.com}

\author{Swapnil Daga}
\affiliation{%
  \institution{Robotics Research Center\\ KCIS, IIIT Hyderabad, India}
}
\email{swapnil.daga@research.iiit.ac.in}

\author{Nahas Pareekutty}
\affiliation{%
  \institution{Robotics Research Center\\ KCIS, IIIT Hyderabad, India}
}
\email{nahas.p@research.iiit.ac.in}

\author{K. Madhava Krishna}
\affiliation{%
  \institution{Robotics Research Center\\ KCIS, IIIT Hyderabad, India}
}
\email{mkrishna@iiit.ac.in}

\author{Balaraman Ravindran}
\affiliation{%
  \institution{Dept. of Computer Science, \\ Indian Institute of Technology}
  \city{Madras, India}
}
\email{ravi@cse.iitm.ac.in}

\author{Brojeshwar Bhowmick}
\affiliation{%
  \institution{Embedded Systems \& Robotics\\TCS Research \& Innovation}
  \city{Kolkata, India}
}
\email{b.bhowmick@tcs.com}


\renewcommand{\shortauthors}{Prasad et al.}
\renewcommand{\shorttitle}{Learning to Prevent Monocular SLAM Failure using Reinforcement \\Learning}
\begin{abstract}
Monocular SLAM refers to using a single camera to estimate robot ego motion while building a map of the environment. While Monocular SLAM is a well studied problem, automating Monocular SLAM by integrating it with trajectory planning frameworks is particularly challenging. This paper presents a novel formulation based on Reinforcement Learning (RL) that generates fail safe trajectories wherein the SLAM generated outputs do not deviate largely from their true values. Quintessentially, the RL framework successfully learns the otherwise complex relation between perceptual inputs and motor actions and uses this knowledge to generate trajectories that do not cause failure of SLAM. We show systematically in simulations how the quality of the SLAM dramatically improves when trajectories are computed using RL. Our method scales effectively across Monocular SLAM frameworks in both simulation and in real world experiments with a mobile robot.
\par
\end{abstract}

%
%


\copyrightyear{2018}
\acmYear{2018}
\setcopyright{acmcopyright}
\acmConference[ICVGIP 2018]{11th Indian Conference on Computer Vision, Graphics and Image Processing}{December 18--22, 2018}{Hyderabad, India}
\acmBooktitle{11th Indian Conference on Computer Vision, Graphics and Image Processing (ICVGIP 2018), December 18--22, 2018, Hyderabad, India}
\acmPrice{15.00}
\acmDOI{10.1145/3293353.3293400}
\acmISBN{978-1-4503-6615-1/18/12}
\maketitle

\let\thefootnote\relax\footnotetext{\textbf{Authors Vignesh and Karmesh contributed equally.}}

\section{Introduction}

Simultaneous Planning Localization and Mapping (SPLAM) or Active SLAM has been a popular area of research over the years. The main theme is to compute a set of control actions for the mobile robot such that either the uncertainty of the robot or the combined uncertainty of robot and map states are bounded. Traditionally, SPLAM frameworks have been approached through either Model Predictive Control \cite{leung2006active,leung2006planning} or Information Gain \cite{kollar2008efficient} paradigms. In recent times, Belief Space planning paradigms have tended to compute the control law in continuous domain \cite{indelman2015planning,charrow2015information} taking a leaf out of SLAM frameworks that model it as a least squares problem \cite{kaess2008isam}. Planning under uncertainty using monocular vision by considering scene texture in addition to the geometric structure of the scene has also shown to improve the localization uncertainty \cite{costante2016perception}. \par

\begin{figure}[h]
    \begin{tabular}{p{3.9cm}p{3.9cm}}
	\vspace{-1.5em}
	\subfloat[]{
	\includegraphics[width=3.9cm, height=4cm]{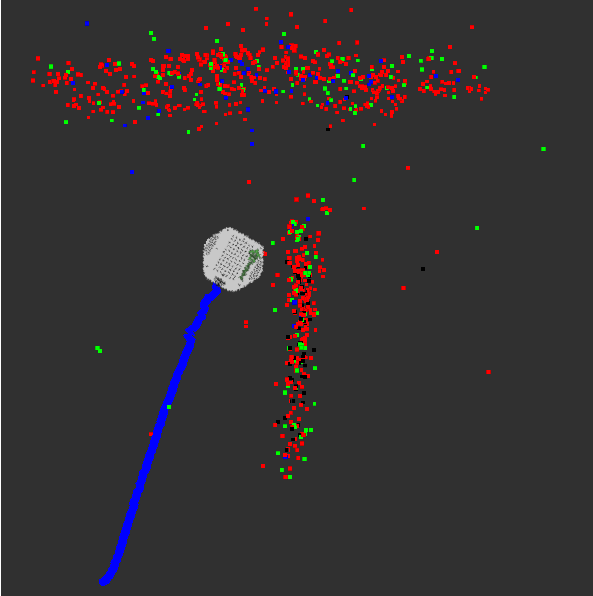}
	\label{teaser_rviz_naive}
	} &
	\vspace{-1.5em}
	\subfloat[]{
	\includegraphics[width=3.9cm, height=4cm]{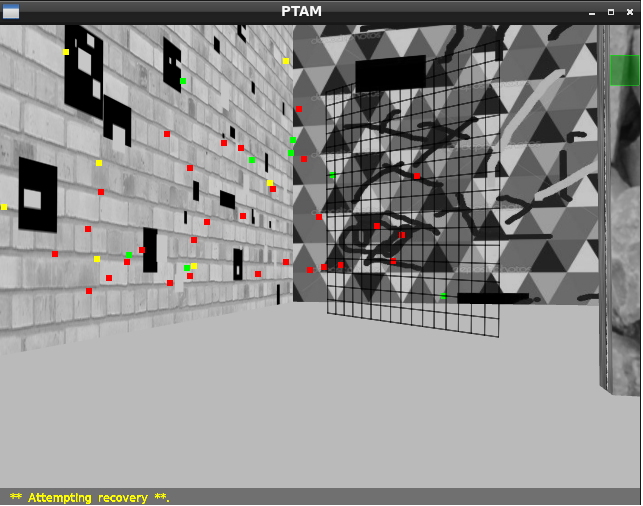}
	\label{teaser_ptam_naive}
	}\\
	\vspace{-1.6em}
	\subfloat[]{
	\includegraphics[width=3.9cm, height=4cm]{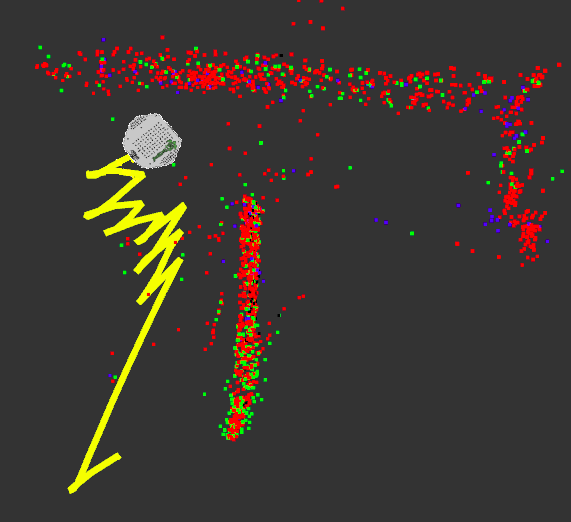}
	\label{teaser_rviz_rl}
	} &
	\vspace{-1.6em}
	\subfloat[]{
	\includegraphics[width=3.9cm, height=4cm]{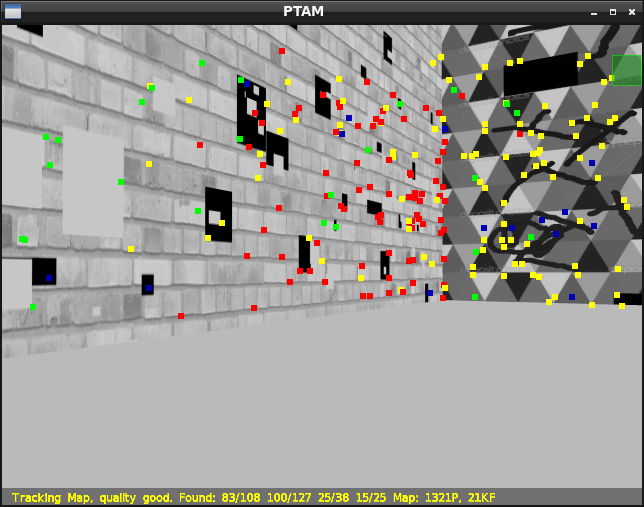}
	\label{teaser_ptam_rl}
	} \\
	\end{tabular}
	\vspace{-1.3em}
	\caption{\small{Robot Trajectory estimate and Map (a) with a naive uninformed planner and (c) with the RL-based planner. The corresponding Monocular SLAM(PTAM) feedback is shown in (b) and (d) respectively.}}
	\label{teaser}
	\vspace{-2em}
\end{figure} 

However, all the above works assume the availability of immediate range data, map uncertainty estimates or dense depth maps of the scene. In contrast, Monocular SLAM gives sparse depth estimates, which are highly inaccurate in texture-less or low texture scenes. These cases usually cause a failure in the feature tracking and consequently in SLAM itself.  Moreover, planar scenes and in-place rotations lead to degeneracies in camera pose estimates. These combined with the non-linear nature of the camera projection operation lead to complexities in Monocular SLAM estimates. 

Literature seems sparse on SPLAM in a monocular setting. A Next-Best-View (NBV) approach \cite{mostegel2014active}, based on estimated measures of the localization quality, showed good results for goal based trajectory planning as well as strategic exploration on Micro Aerial Vehicles (MAVs). Nonetheless it is shown subsequently that the current method evaluates better on various maps in terms of SLAM failures.
Autonomous navigation with a downward facing camera for Micro Helicopters was shown in \cite{weiss2011monocular}. In our case, we try to autonomously navigate a non-holonomic robots with a front facing camera. This is more constrained than the above mentioned works where the use of MAVs allow smooth lateral motions thus providing wide baselines to improve the quality of Monocular SLAM, whereas in our case we use a ground robot restricted to move either forward or backward. \par

Prasad \textit{et al.}\cite{prasad2017data} approach the problem using an Apprenticeship Learning formulation\cite{abbeel2004apprenticeship}. However their approach is time consuming during training, where they evaluate the learnt reward at each iteration and the efficacy of this formulation for larger scale maps is also unproven.\par

In contrast, this paper formulates the SPLAM problem through a Reinforcement Learning paradigm in a map agnostic fashion. Our approach is trained on a single map and is applied on a variety of maps without any re-training/fine-tuning. Instead of learning outright an optimal policy, the current framework learns actions detrimental to SLAM using Temporal Difference Learning methods \cite{sutton1998reinforcement}. We learn an optimal action-value function with respect to SLAM failure which is used to filter out potentially unsafe actions. This learnt optimal Action Filter is seamlessly used across a variety of maps showing significant reduction in Monocular SLAM failures, which are mainly characterized by tracking failures or due to large errors in the pose priors used for estimation. \par

By using RL in an episodic manner, the result of each failure is propagated back to earlier states over multiple episodes, making the predictions of /submit/2514950/addfilesfuture failures more likely in the earlier stages once the learning converges. Analysis shows significant reduction in SLAM failures when the learned Q values are used to filter actions in trajectories generated by routine methods such as sampling based planners \cite{lavalle2001randomized} or trajectory optimization routines \cite{gopalakrishnan2014time} \cite{bertram2013trajectory}. Qualitative and quantitative analysis and comparisons with supervised learning approaches, 3D point overlap maximization techniques, and predicting Localization Quality estimates, described in \cite{mostegel2014active}, showcase the superior performance of the proposed method. The effectiveness of RL based navigation is also demonstrated with different feature based Monocular SLAM systems, in both simulated and real world indoor scenes with a ground robot. We further show the advantage of our algorithm when augmented with relocalization after SLAM failure and with multi-modal sensor fusion with SLAM, leading to almost 100\% prevention of SLAM failure. \par 

The keynote is that almost all previous Monocular SLAM results have been shown with a handheld camera or on a robot that is teleoperated by a user. In such scenarios, human intuition and experience is often responsible for the success of SLAM. For example, in the case of handheld cameras it is common for the user to repeatedly scan the same area to improve the quality of results. In this paper the robot negotiates turns through a sequence of back and forth motions, tacitly avoiding a single shot acute turn or bend. RL based methods are amenable to learn such experience and intuition and we dovetail it to a RL based planner forming the essential theme of this work. This is precisely captured in the trajectory of Fig. \ref{teaser_rviz_rl}, where a sharp turn is decomposed into a sequence of back and forth maneuvers that has been learned by RL. Such a maneuver provides for SLAM stability by attenuating a sharp turn and allowing the vehicle to look at an area for a longer time. This helps to ensure persistence of mapped points over multiple frames thereby improving the accuracy of the estimates of both the pose and the sparse map. \par

\begin{figure*}[h!]
\centering
\includegraphics[width=\textwidth]{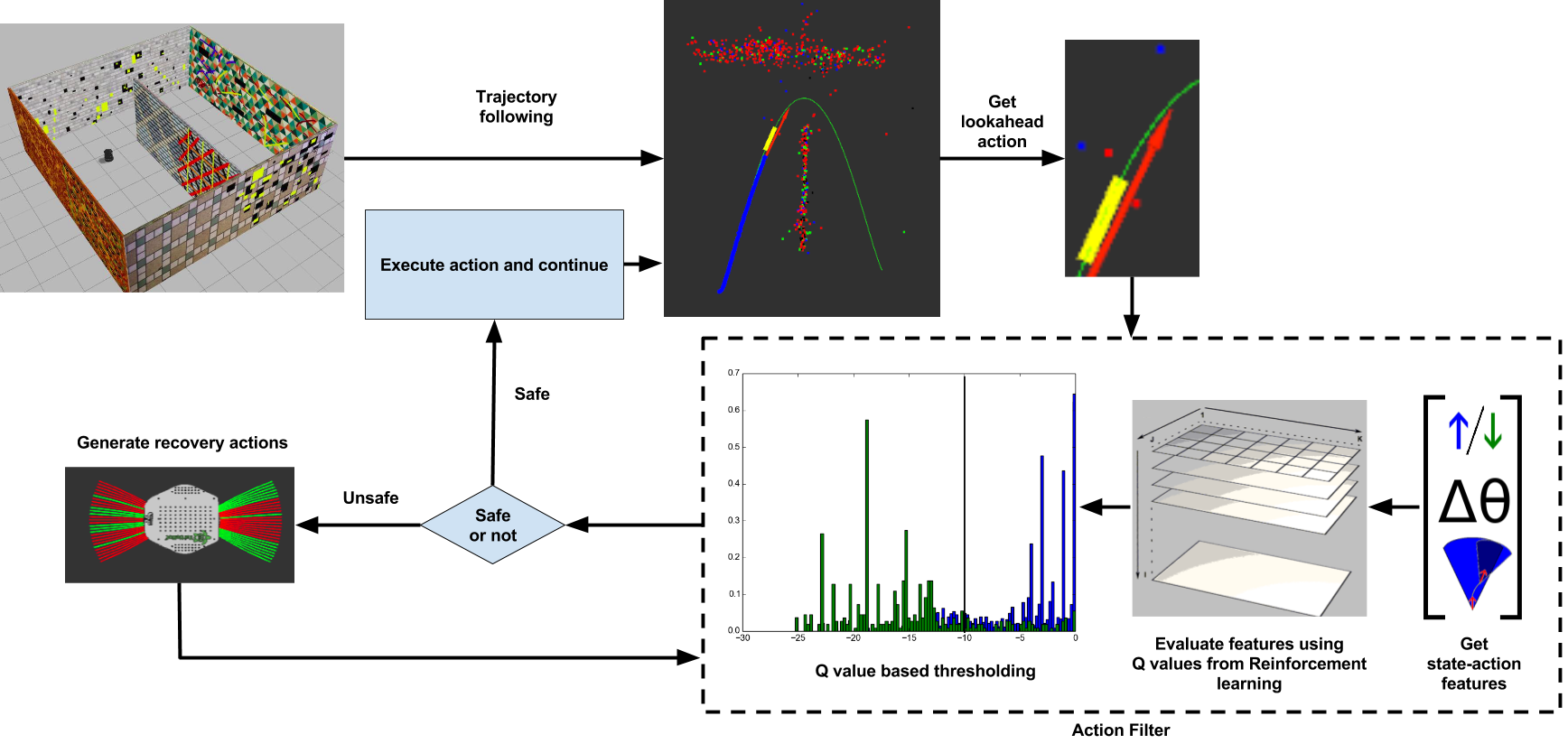}
\caption{\small{Overview of the RL based Navigational Planner. Once a global trajectory is generated by an underlying planner, as it is executed, the subsequent part of the trajectory (shown with the yellow marker), along with the current set of points are taken as the state-action pair and the Q value for it is calculated. The Q value is checked to see if the motion that is to be executed is safe or not. If its not, then sample recovery actions are generated and among those with a safe Q value (shown in green in the bottom left), the one that best aligns the robot with the global trajectory is executed. After this, a new global trajectory is generated from the new pose and this cycle continues till the robot reaches the goal. }}
\label{navig_planner}
\end{figure*}

\section{\textbf{Background} \label{background}}
\subsection{Monocular SLAM} \label{monoSLAM}

Monocular Simultaneous Localization and Mapping (Monocular SLAM) refers to the method of estimating the camera pose while performing a 3D reconstruction of the environment using only a monocular camera. \textbf{PTAM (Parallel Tracking and Mapping)} \cite{klein2007parallel} was one of the first such SLAM systems.  \textbf{ORB-SLAM} \cite{mur2015orb} is a newer feature based Monocular SLAM that builds on the main ideas of PTAM, using different image features for tracking as compared to PTAM and incorporates real time loop closure as well. \textbf{LSD-SLAM} \cite{engel2014lsd} is another state-of-the-art Monocular SLAM system which works with image intensities and creates a semi-dense map of the environment. A survey of various Monocular SLAM methods can be found in \cite{younes2017keyframe}.\par
While these methods have shown effective SLAM capabilities, they are susceptible to erroneous pose estimates due to insufficient tracking of mapped points or motion induced errors such as large or in-place rotations. The latter can be mitigated to an extent by restricting robot motion, but this may lead to inefficiencies in navigation. A promising solution would therefore be based on automatically learning robot behavior that navigates while keeping SLAM failures to a minimum. Considering the requirements of self-learning and adaptability, Reinforcement Learning forms a good candidate for such a solution. \par

\subsection{Reinforcement Learning} \label{RL}
\noindent

Reinforcement Learning (RL) is a learning method based on Markov Decision Processes (MDPs) where actions are performed based on the current state of the system and rewards are obtained accordingly. The correspondence between states and the actions performed from each state is determined by the policy, $\pi$. The aim of the decision process is to optimize the policy, which in turn is performed by maximizing the return from each state. In formal notations, an MDP is a tuple $(S, A, T, \gamma, R)$, where 
\begin{itemize}
\item $S$ is a set of states.
\item $A$  is a set of actions.
\item $T$ is a set of state transition probabilities.
\item $\gamma$ is the discount factor. $\gamma \in [0,1]$
\item $R$ is the reward function. $R : S \times A \rightarrow \mathbb{R}$
\end{itemize}

The effectiveness of an action is defined by the Q value of the action and the state from which it is performed i.e. state-action pair, denoted by $Q(s, a)$. Once learned effectively, it takes into account not just the immediate reward but also the future discounted rewards of the transitions that follow.


In this paper, we use Q-Learning \cite{watkins1989learning,watkins1992q} for learning navigational strategies. Q-Learning is a widely used Temporal Difference prediction method for control problems. It learns optimal Q values for a given MDP. Starting with an initial estimate, these values are updated incrementally through learning episodes or samples, in a way similar to dynamic programming. The update for performing an action \textit{a} from a state \textit{s}, reaching a new state \textit{s'} and obtaining a reward \textit{r} is as follows:
\begin{equation} \label{q-learning-eq}
Q(s,a) \leftarrow Q(s,a) + \alpha [ r + \gamma\max_{a'}Q(s',a') - Q(s,a)]
\end{equation}
where \textit{a'} is the action that would be performed from the next state \textit{s'}, $\alpha$ is the learning rate and $\gamma$ is the discount factor. Once the Q values are learnt, the optimal action is the one which maximizes the Q value from the current state.

\section{\textbf{Preventing SLAM Failure using RL}}

The motivation for this paper is to devise a navigational methodology for a robot that prevents Monocular SLAM failure. This is ensured by using RL to learn the relationship between robot actions and SLAM failure. 
The robot essentially learns to avoid potentially unsafe movements that can lead to SLAM failure. Fig. \ref{navig_planner} shows a flowchart of the complete navigational planner, which is explained below in detail. \par

\subsection{Navigation Overview} \label{planner}

Initially, a trajectory is planned from the current location to a goal, as depicted by the green curve in the middle image of Fig. \ref{navig_planner}. Goals are selected at frontier locations of the map with enough overlap with already mapped points to prevent SLAM failure. \par



Once the robot starts moving, the immediate subsequent part of the trajectory, which is shown in yellow in the same image of Fig. \ref{navig_planner}, is parametrized into a state-action pair, details of which can be found in Sec. \ref{learning}. An Action Filter evaluates whether this action will cause a SLAM failure or not. This is done by comparing the Q value of the corresponding state-action pair against a safety threshold. This is shown in the bottom right image of Fig. \ref{navig_planner}.  
If the Q value falls below the threshold, i.e. if an unsafe motion is predicted, the robot is stopped and potential recovery goals are generated and checked by the Action Filter. The recovery actions are trajectories to points at a distance of 1m form the current position in both forward and backward directions with $0^\circ-30^\circ$ deviations on either side (left/right). This can be seen in the bottom left image of Fig. \ref{navig_planner}, where the green trajectories are the safe trajectories and the red ones are unsafe. We restrict the angle to $30^\circ$ as angles larger than this, in such a small distance, would be detrimental for SLAM. We choose these actions at a distance of 1m as points further away would lead to lesser overlap between images and more importantly, in 3D scene points.\par

Among those predicted as safe, the pose that best aligns it with the global path is chosen for execution. If the chosen pose is too close to the unsafe ones, it is discarded and the pose with the best Q value is chosen. Once this recovery trajectory is executed, a new path is generated from the new location to the goal. This process is cyclically repeated until the goal is reached. \par

\subsection{Learning to prevent SLAM failure using RL} \label{learning}
To effectively design the Action Filter, the quality of an action should be primarily dependent on the probability of it causing a SLAM failure while being independent of the map(s) being used for learning.  \par

The objective of the RL method is to evaluate the quality of an action performed from a state with respect to the SLAM pose estimate. The model was trained in a simulated environment, by generating sample paths from the robot's current pose. 
We use Q-Learning for performing the learning, which takes as input a state-action pair $(s,a)$. This leads us to use the SLAM related outputs from the current and the next pose. Our state space comprises of the number of common map points seen between the subsequent poses, $N_{3D}$ while the action space includes the direction of motion (forward or backward), represented as $\eta$ and the angle change, $\Delta\theta$ between the current and the next pose. \par

We observed an increasing trend in the frequency of failures as the change in heading angle increased. This causes the pose estimate of SLAM to deteriorate, which leads to SLAM failure. Therefore we restrict our learning to a maximum of 30$^\circ$. Coming to $N_{3D}$, we keep a cap of 600 points as anything more would anyway not contribute to SLAM failure or cause deterioration of SLAM estimates. \par

We use a lookup table representation for the Q function, by discretizing the parametrized state-action pair. Details of this discretization are given in table \ref{discrete}. \par

\small
\begin{table}[h]
\caption{Discretization of State-Action pairs}
\vspace{-0.5em}
\label{discrete}
\begin{center}
\begin{tabular}{p{2cm} p{3cm} p{2cm}}
\hline
Parameter & Range & No. of discretized values \\
\hline
$\eta$ & forward or backward & 2 \\
$\Delta\theta$ & 0$^\circ$ to 30$^\circ$ & 20 \\
$N_{3D}$ & 0 - 600 & 20 \\
\hline
\end{tabular}
\end{center}
\end{table}
\normalsize

The Q values were learnt by performing Q-Learning with a weighted combination of the above parameters as the reward function. The reward function weights are chosen by making a few assumptions about the way Monocular SLAM Systems behave. Other than observing the change in heading angle and overlap of points, the status of SLAM after performing an action is also used, which we denote as $\Phi$. This is a Boolean value which tells us if SLAM failure occurs due to a particular action. This is an observable entity obtained from Monocular SLAM used during the training phase only. \par 
The reward function of our model is a weighted combination of these parameters given in Eq. \ref{getpenalty}.
\begin{equation} \label{getpenalty}
R(s,a) = \omega_{3D}N_{3D} + \omega_{\theta}\Delta \theta + \omega_{\Phi}\Phi + \omega_0
\end{equation}
The weights were assigned based on the extent to which these parameters affect Monocular SLAM. 
\begin{itemize}
    \item The weight $\omega_{3D}$ is given a positive value of 0.0167 (with the maximum value of $N_{3D}$ being 600, it allows a maximum reward of +10) as the larger the overlap between frames, the better it is for Monocular SLAM.
    \item For $\omega_\theta$, a negative weight of -0.1 is given(with the lowest reward being -3) as larger angle change degrades the quality of SLAM.
    \item Since our main aim is to prevent SLAM failure a relatively large value of -10 is given to $\omega_{\Phi}$. 
    \item The value of the bias term, $\omega_0$ is set to -10 so that the overall reward is kept negative.
    \item There is no separate reward for successfully reaching a goal state. Our aim is not to optimally reach the goal location, but to prevent SLAM failure along the way.
\end{itemize}


\section{\textbf{Experimentation}}
\subsection{Training}
Gazebo \cite{koenig2004design} is a framework for robot simulations and it reproduces dynamic environments like those a robot would encounter in the real world. Indoor maps were created in Gazebo to simulate real-life environments.

\begin{figure}[h]
	\centering		
    \includegraphics[width=0.4\textwidth]{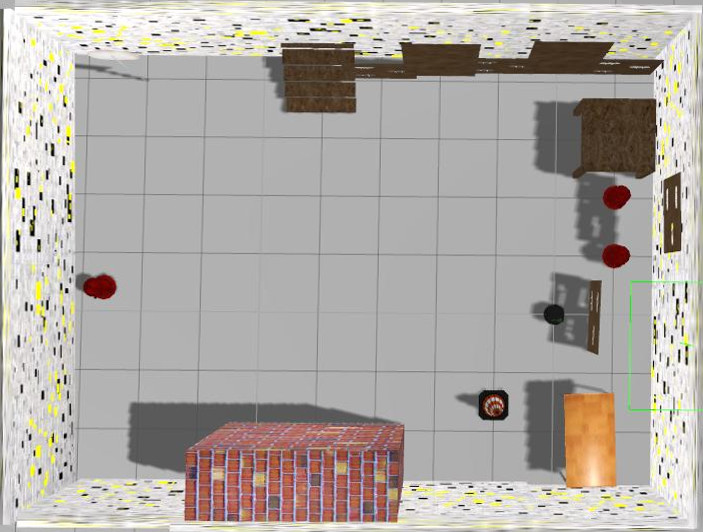}
	\caption{\small{A map of the simulated training environment in Gazebo.}} 
	\label{gazebo_map}
\end{figure} 

\begin{figure}[h!]
\centering	
	\includegraphics[width=0.4\textwidth]{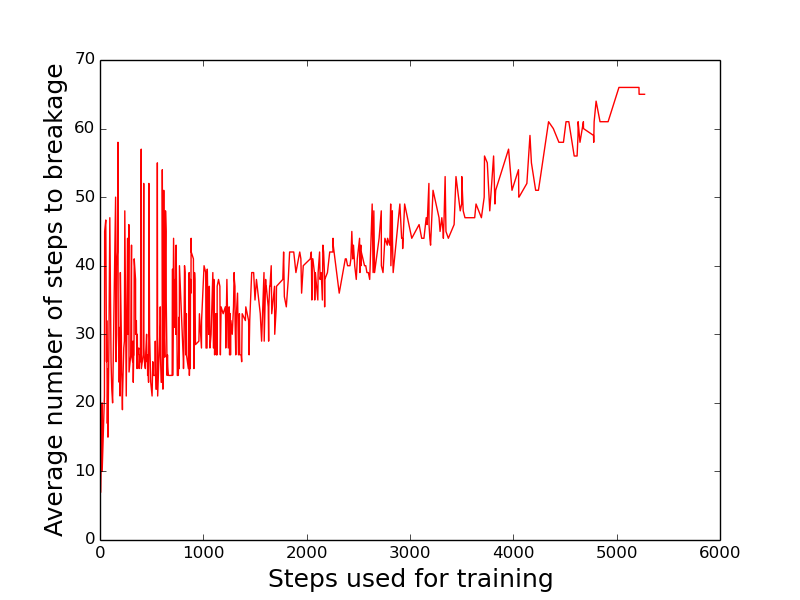}
	\caption{\small{The average steps to taken by the robot until SLAM failure in the training phase.}}
	\label{break_q_fall}
\end{figure}

Training is carried out in a training map (7mx10m) using PTAM, shown in Fig. \ref{gazebo_map}. Navigation during episodic learning consisted of random walks with an Action Filter derived from the learnt Q values. An episode is the sequence of steps until SLAM failure. Initially, motions were randomly generated to learn an initial set of Q values. Once the Q values are initialized, we use an incremental $\epsilon$-Greedy policy to reinforce the learnt Q values after each episode using Q-Learning. 
After each set of episodes, the value of $\epsilon$ is incremented by a small amount until it reaches 0.9, as can be seen in Alg. \ref{inc_greedy_train}. Basically, once a basic set of values are learnt, we keep increasing the exploitation ratio in order to further reinforce what is learnt. This can be seen in the increasing trend in the number of steps taken till SLAM failure, as seen in Fig. \ref{break_q_fall}. Initially, due to the high level of randomness, there is a high variance in the output but as the learning progresses, there is an increasing trend in the average number of steps taken till SLAM failure. \par 
One thing to note is that unlike most RL problems where the average reward per step or similar metrics would plateau, our metric, i.e. the average number of steps to SLAM failure, would keep increasing as the learning progresses. Ideally, the robot would never get into a situation where SLAM would fail. One possible way to combat this is to keep a cap on the maximum number of steps per episode. This could possibly help speed the learning as the larger the number of steps go, the greater they get discounted so it would take longer time to propagate its effect to the initial states. However, we defer this exercise for our future work. \par

\begin{algorithm}[h]
	\caption{Incremental $\epsilon$-Greedy Training}\label{inc_greedy_train}
	\begin{algorithmic}
    	\State $\epsilon \gets 0.1$
    	\State Repeat (while $\epsilon < 1$):
    	\begin{algorithmic}
    	    \State Repeat (for \textit{N} episodes):
    	    \begin{algorithmic}
    	        \State Initialize state \textit{s}
        	    \State Repeat (for each step in the episode):
    	        \begin{algorithmic}
    	            \State Choose action \textit{a} at \textit{s} using an $\epsilon$-Greedy policy
    		        \State Execute action \textit{a} and get reward \textit{r} and reach next state \textit{s'}
    		        \State $Q(s,a) \leftarrow Q(s,a) + \alpha [ r + \gamma\max_{a'}Q(s',a') - Q(s,a)]$
    	    	    \State \textit{s} $\gets$ \textit{s'}
    		    \end{algorithmic}
    	    \end{algorithmic}
    	\end{algorithmic}
    	\State $\epsilon \gets \epsilon + 0.1$
	\end{algorithmic}
\end{algorithm}

Once the Q values are learnt, we use it in the Action Filter to classify actions as failing or not. The optimal quality threshold for the Action Filter would precisely classify state-action pairs into SLAM failures/non-failures. The quality threshold was chosen based on the primary objective of minimizing false positives (unsafe actions classified as safe) and a secondary objective of minimizing false negatives (safe actions classified as unsafe). From our experiments, we found a threshold of -10 to be suitable for achieving good performance.

The main takeaway is that the training was performed on a single map and was seamlessly applied on various maps of varying structure and sizes, without the need for retraining. The same Q-values learnt using PTAM were able to be applied directly to ORB-SLAM as well, by just changing the thresholds. This highlights the effectiveness of our map-agnostic approach and also shows that it can be effortlessly applied to different Monocular SLAM algorithms as well.



\subsection{Comparison with Alternative Policies}

In order to get further insight into the effectiveness of our method, we substituted the learnt policy with three alternative policies. One using supervised learning, one based on the common mapped points between subsequent poses, which we defined earlier as $N_{3D}$ and a Next-Best-View method. 

We decided to use support vector machines (SVM) \cite{cortes1995support} over other classifiers like artificial neural networks, decision trees etc. due to better average performance tested using Monte Carlo cross-validation. A SVM with a radial basis function (rbf) kernel, implemented using scikit-learn \cite{pedregosa2011scikit}, is trained to classify trajectories as failing or not. The distance of samples from the decision boundary are used to give a score to the trajectories. The same parameters which were used by RL, are used as inputs for the SVM. \par 

Superior performance is also shown in comparison to directly using $N_{3D}$ to detect failures and choose actions. This proves that feature overlap isn't the only parameter adversely affecting Monocular SLAM and other parameters need to be considered to effectively navigate in such a setting. \par

Comparison with a Next-Best-View formulation \cite{mostegel2014active}, based on estimating the localization quality is also shown. This is done based on the Geometric Point Quality and the Point Recognition Probability of the mapped points. The Geometric Point Quality penalizes points with low triangulation angles, as this increases localization uncertainty. The Point Recognition Probability gives an estimate of the recognizability of a point based on the viewing angle and the scale at which the point was originally found. This provides a faster evaluation of map uncertainty as compared to calculating covariance matrices. A detailed explanation can be found in \cite{mostegel2014active}.\par

\section{\textbf{Result of RL based Navigation}}

Experiments were conducted by navigating a Turtlebot on different maps in Gazebo environment. For the physical implementation a Pioneer 3-DX was used. In all the experiments, the color image output of a Microsoft Kinect was used as the video source. Experiments were carried out on a laptop with Intel Core i7-7700U 2.80GHz CPU running Ubuntu 16.04 using Robot Operating System (ROS) \cite{quigley2009ros} for controlling the robot and performing SLAM.\par 

\subsection{Steps till SLAM Failure}

\begin{figure}[!h]
\centering
	\includegraphics[width=8cm, height=5cm]{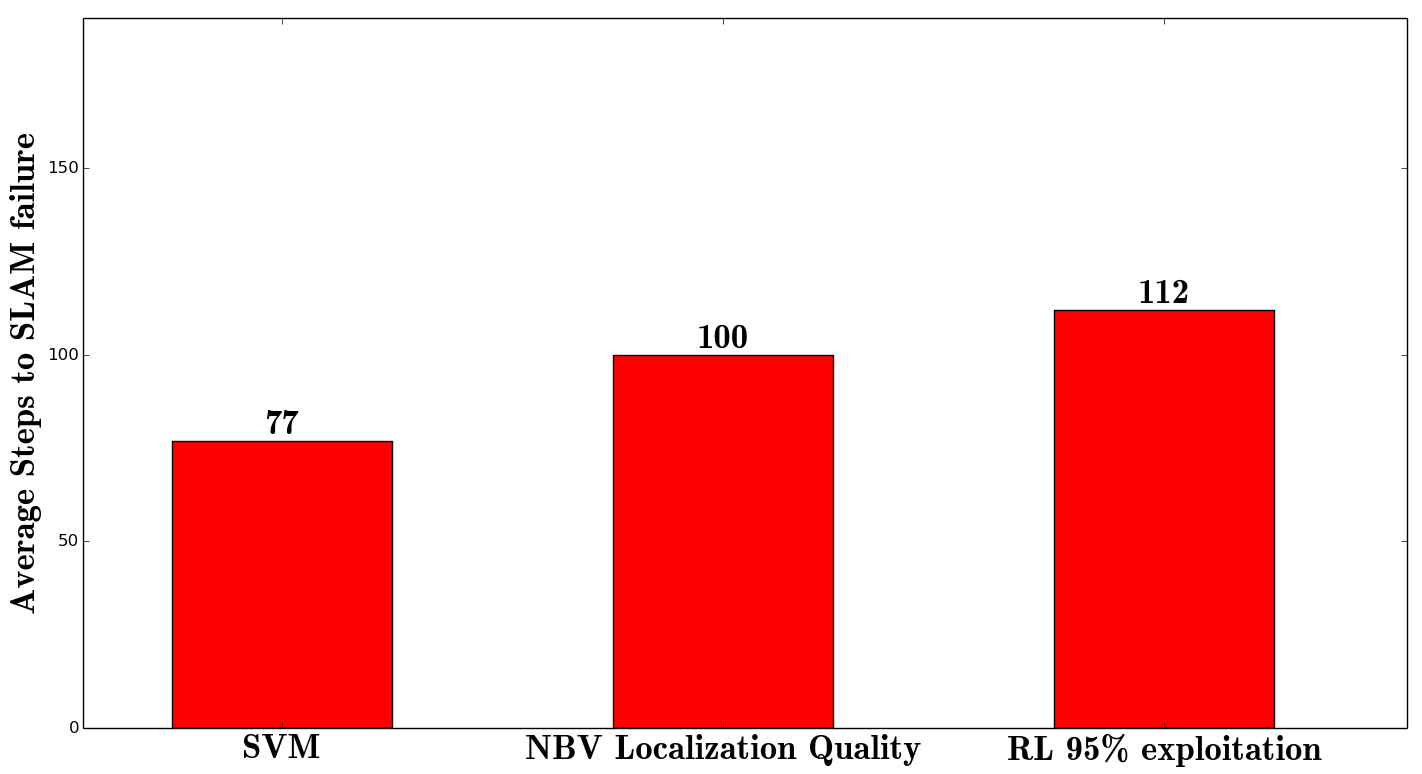}
	\caption{\small{Comparison of average steps to SLAM failure using SVM, Localization Quality based Next-best-view (NBV)\cite{mostegel2014active} and RL. The results are averaged over 10 episodes each.}}
	\label{steps_to_break}
\end{figure}

To evaluate the performance of our method, we compare the average steps executed by the robot from SLAM initialization till SLAM failure. By step, we mean a trajectory to a distance of 1m from the current position in both forward and backward directions with $0^\circ-30^\circ$ deviations on either side (left/right), similar to the recovery action generation. This is evaluated in the training map, shown in Fig. \ref{gazebo_map}. The robot starts at an arbitrary location in the map and performs actions based on the policy to be evaluated.\par

As seen in Fig. \ref{steps_to_break}, our policy performs better than all the other policies considered. Since the RL model is episodically trained, the result of each failure gets propagated back over episodes, making the predictions of future failures more likely in the earlier stages as the learning goes on. Hence the predictions are more accurate than an SVM, which was trained by just a single label for each step (failing or not).

\begin{figure*}[!h]
	\centering
	\begin{tabular}{>{\centering\arraybackslash}m{2cm} >{\centering\arraybackslash}m{3.5cm} >{\centering\arraybackslash}m{3.5cm} >{\centering\arraybackslash}m{3.5cm} >{\centering\arraybackslash}m{3.5cm}}
    & \textbf{Map 1} & \textbf{Map 2} &\textbf{Map 3} &\textbf{Map 4} \\
   \textbf{Environment} &
   \subfloat[]{
	\includegraphics[width=3.5cm, height=3.5cm]{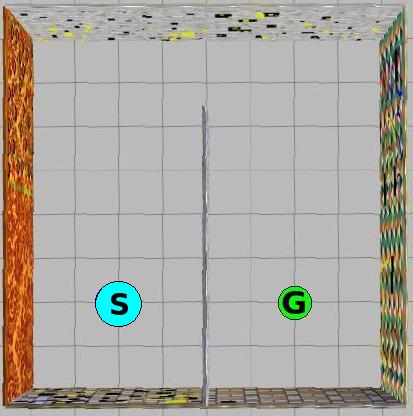}
	}\label{map1} &
 	\subfloat[]{
 	\includegraphics[width=3.5cm, height=3.5cm]{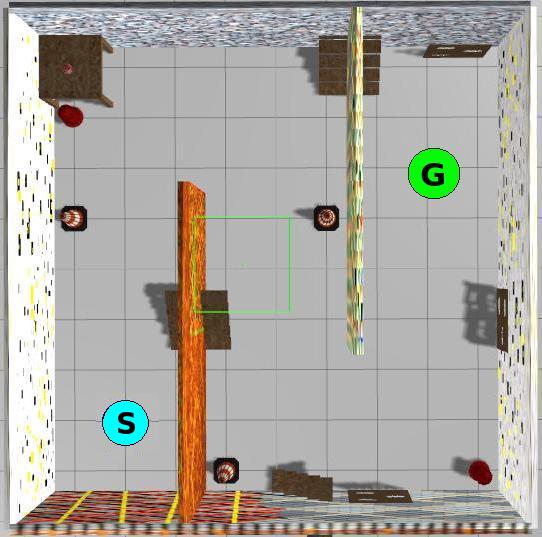}
	}\label{map4} &
 	\subfloat[]{
 	\includegraphics[width=3.5cm, height=3.5cm]{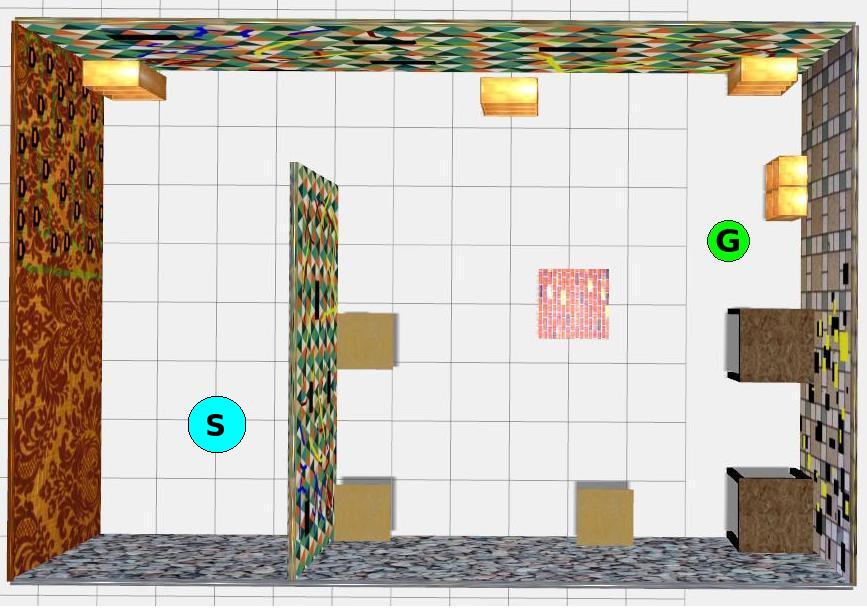}
 	}\label{map5} &
    \subfloat[]{
	\includegraphics[width=3.5cm, height=3.5cm]{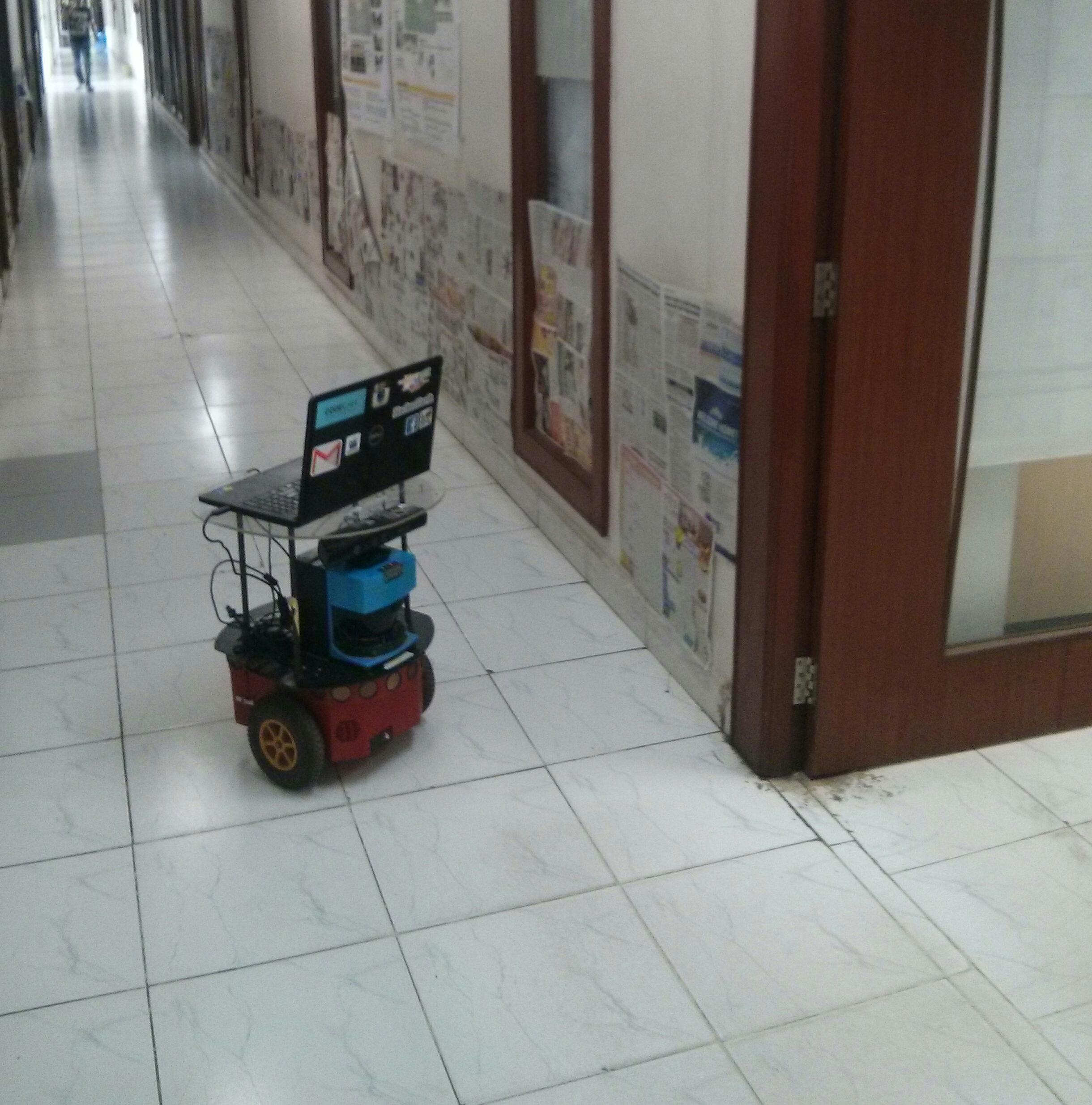}
	}\label{map6} \\
  	\vspace{-1em}
 	\textbf{RL Trajectories with PTAM} &
    \vspace{-1em}
	\subfloat[]{
	\includegraphics[width=3.5cm, height=3.5cm]{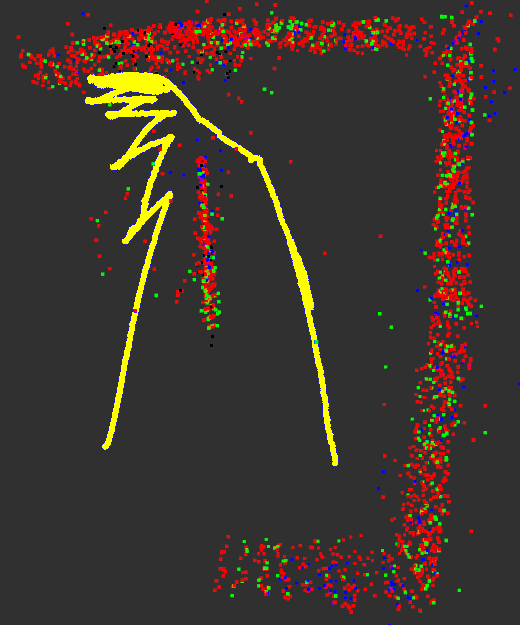}
	}\label{maprun1} &
\vspace{-1em}
 	\subfloat[]{
 	\includegraphics[width=3.5cm, height=3.5cm]{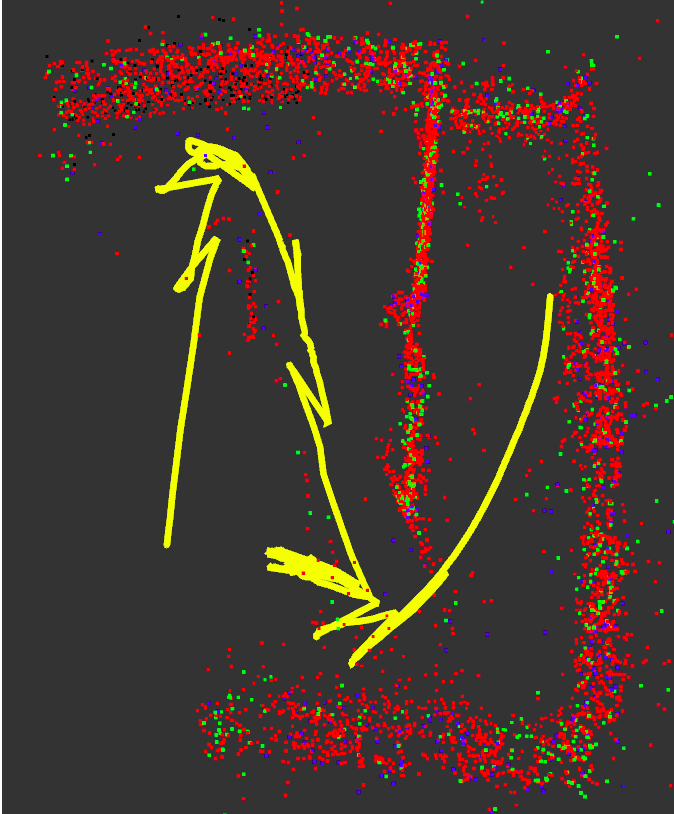}
 	}\label{maprun4} &
    \vspace{-1em}
 	\subfloat[]{
 	\includegraphics[width=3.5cm, height=3.5cm]{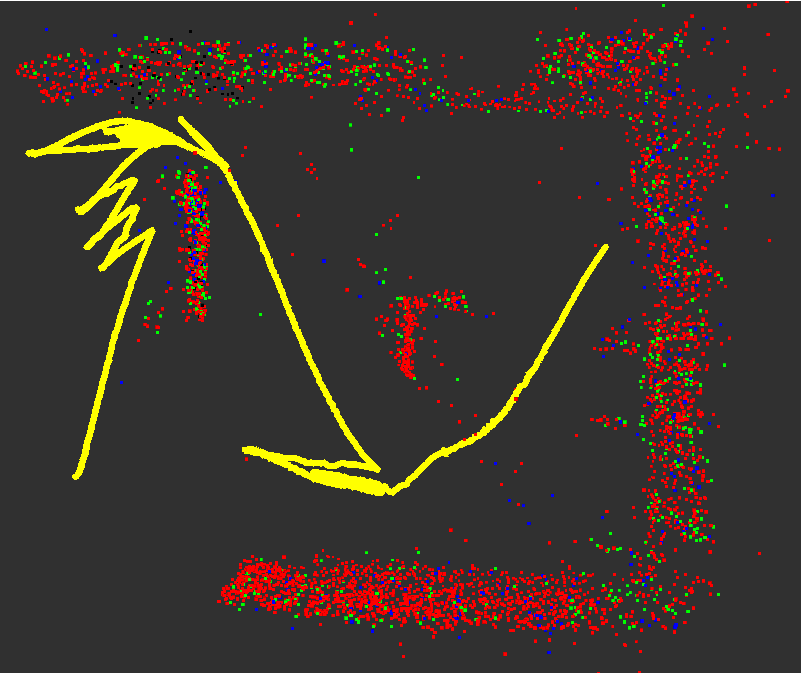}
 	}\label{maprun5} &
    \vspace{-1em}
   \subfloat[]{
	\includegraphics[width=3.5cm, height=3.5cm]{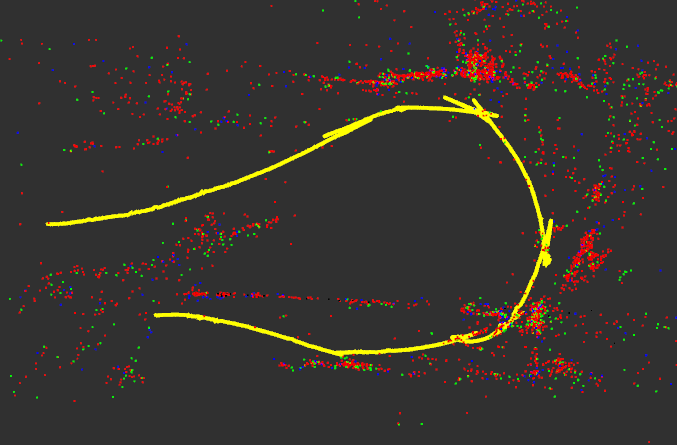}
	}\label{map6_pc_rl} \\
    \vspace{-1em}
    \textbf{RL Trajectories with ORB SLAM} &
    \vspace{-1em}
 	\subfloat[]{
 	\includegraphics[width=3.5cm, height=3.5cm]{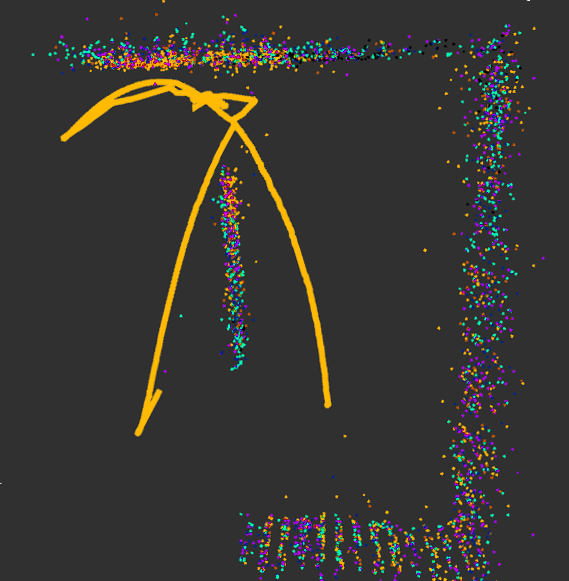}
 	}\label{maporb1} &
    \vspace{-1em}
 	\subfloat[]{
 	\includegraphics[width=3.5cm, height=3.5cm]{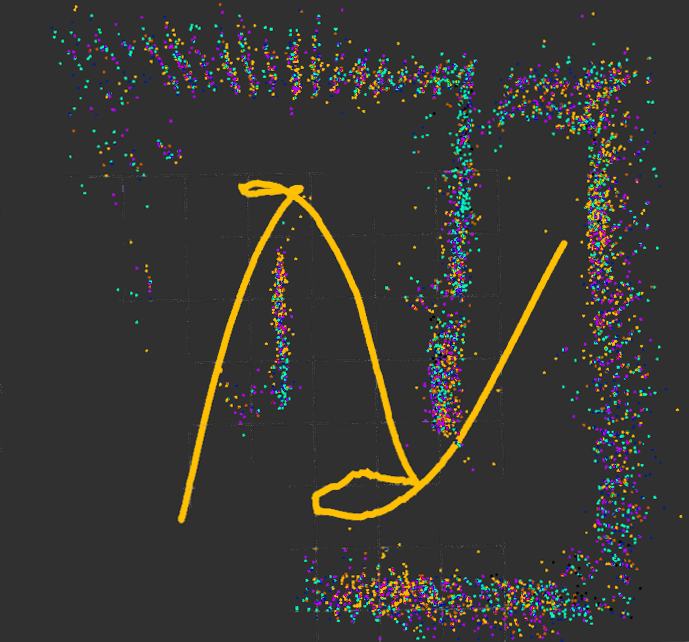}
 	}\label{maporb4} &
    \vspace{-1em}
 	\subfloat[]{
 	\includegraphics[width=3.5cm, height=3.5cm]{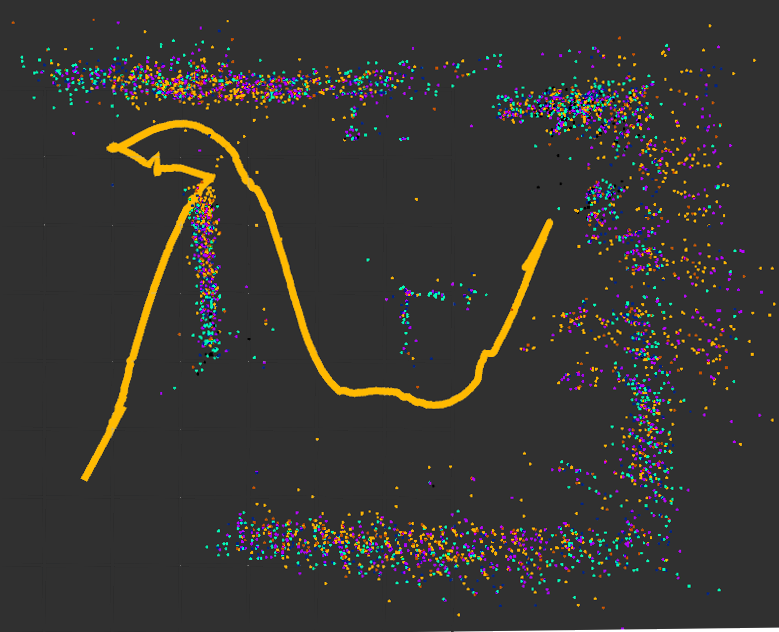}
 	}\label{maporb5} &
    \vspace{-1em}
   \subfloat[]{
	\includegraphics[width=3.5cm, height=3.5cm]{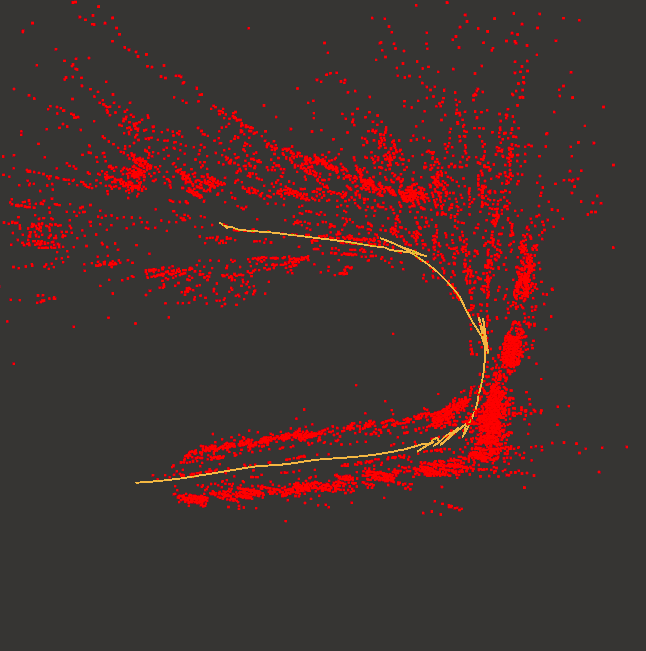}
	}\label{orb_real_run} \\

	\end{tabular}
	\caption{\small{Results of Goal based trajectory navigation using Bernstein Curves. The Gazebo map shows the start and goal locations with blue and green circles, marked as "S" and "G" respectively. The second and third row shows the Trajectory and Map estimates using PTAM and ORB SLAM with the RL based planner respectively.}}
	\label{gaz_results}
	\vspace{-2em}
\end{figure*}

\subsection{Goal Based Navigation}
In order to show the robustness of our approach over different trajectory generators, we combine our RL based approach with Bernstein curves \cite{gopalakrishnan2014time} for trajectory generation and with the ROS Navigation Stack, in which navigation carried out in two levels where Informed RRT* \cite{barfoot2014informed} behaves as the global planner and Timed Elastic Band (TEB) \cite{bertram2013trajectory} performs the role of the trajectory generator. We show results both without (naive) and with RL, to show that an informed approach using RL is more effective than an uninformed or naive path planner.\par

\subsubsection{Bernstein based planning}

After training the robot with PTAM, we test our approach in both simulation and real world. Navigation is performed using Bernstein Curves for path planning, with the robot moving from a predefined start to goal point on different maps using PTAM or ORB SLAM. The results obtained from the RL based planner are also compared against those obtained by using SVM, $N_{3D}$ and NBV\cite{mostegel2014active} based policies. Table \ref{results_table1} summarizes the results of our experiments with each row showing the number of successful runs with the total number of runs given in the brackets. Maps 1-3 are of experiments done in Gazebo and map 4 is of a real environment run. 


The last column of Table \ref{results_table1} shows the number of successful exploration run in a real world environment with Bernstein curves. The robot was made to enter a room from the adjoining corridor and explore the area in the room. The last row of table \ref{results_table1}, corresponding to Map 4, shows the results for this run. \par

\small
\begin{table}[h]
\vspace{-0.5em}
\caption{Results for Goal Based Trajectory Planning (on Map 1-4) using Bernstein with PTAM and ORB SLAM in terms of no. of successful runs. The no. of runs on each map are given in brackets. These results can be seen in Fig. \ref{gaz_results}}
\label{results_table1}
\begin{center}

\begin{tabular}{|c|c|c|c|c|c|}
\hline
&&&&&\\

\textbf{SLAM} & \textbf{Policy} & Map 1 & Map 2 & Map 3 & Map 4\\
&&(10 runs)&(10 runs)&(10 runs)&(5 runs)\\
\hline
PTAM & Naive & 2 & 0 & 0 & 1 \\
PTAM & SVM & 4 & 3 & 3 & -\\
PTAM & $N_{3D}$ & 6 & 0 & 0 & - \\
PTAM & NBV\cite{mostegel2014active} & 7 & 2 & 2 & -\\
PTAM & RL & \textbf{9} & \textbf{8} & \textbf{6} & \textbf{4}\\
\hline
ORB-SLAM & Naive & 3 & 0 & 1 & - \\
ORB-SLAM & RL & \textbf{9} & \textbf{8} & \textbf{6} & \textbf{5}\\
\hline
\end{tabular}
\vspace{-1em}
\end{center}
\end{table}
\normalsize

\subsubsection{Navigation Stack based planning}
\begin{figure}[t]
	\centering
	\subfloat[]{
	\includegraphics[width=3.5cm, height=3.5cm]{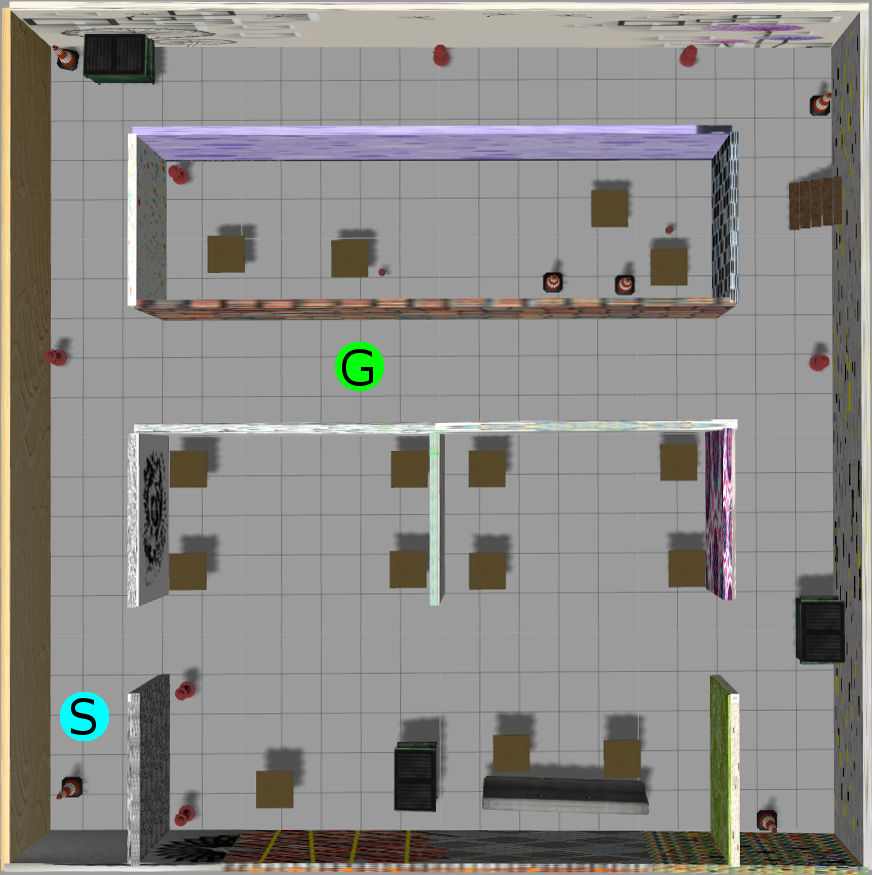}
	}
      \subfloat[]{
	\includegraphics[width=3.5cm, height=3.5cm]{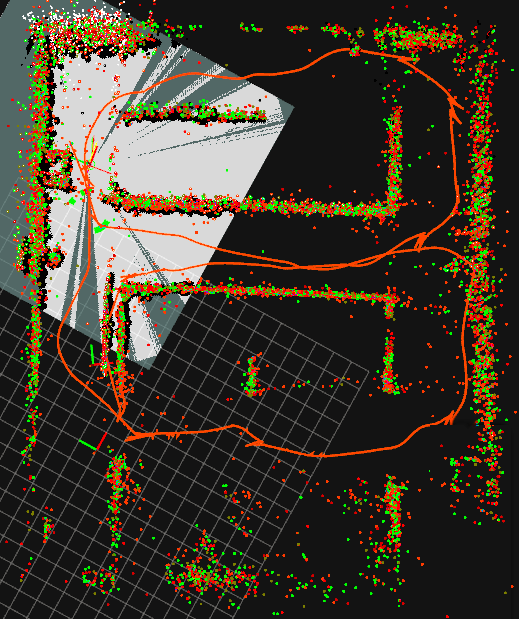} 
    }\\
    \vspace{-1em}
	\subfloat[]{
	\includegraphics[width=3.5cm, height=3.5cm]{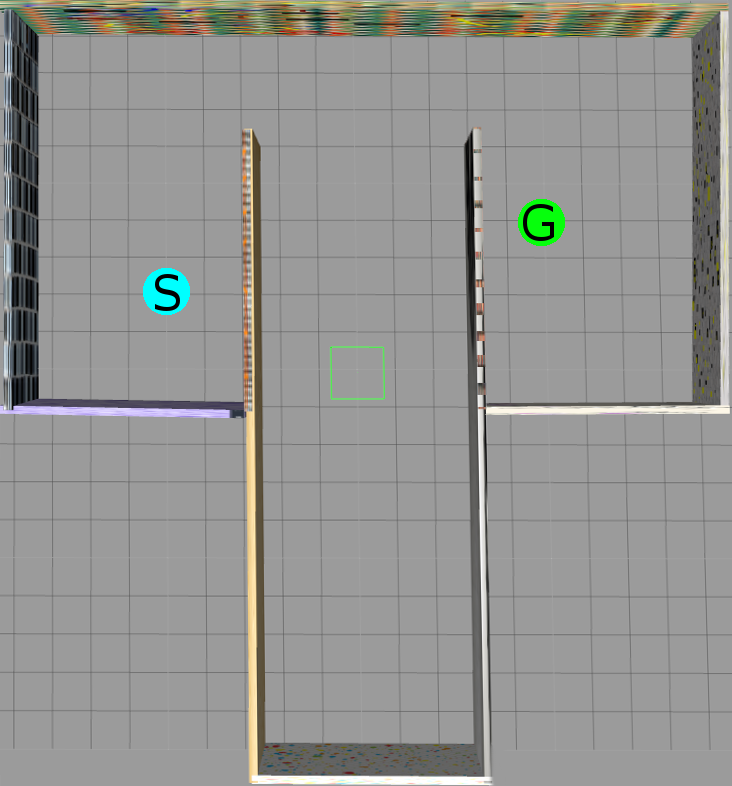}
	}
    \subfloat[]{
	\includegraphics[width=3.5cm, height=3.5cm]{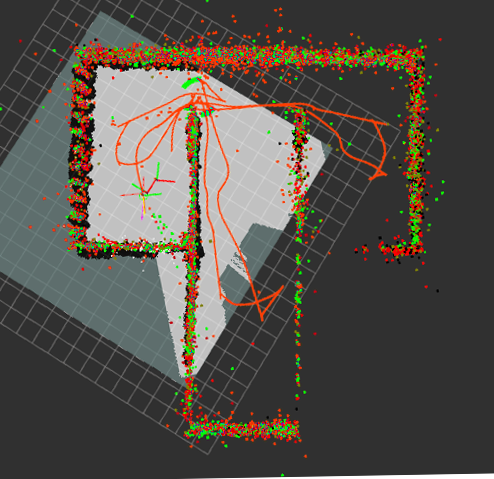} 
    }\\
    \vspace{-1em}
    \subfloat[]{
	\includegraphics[width=3.5cm, height=3.5cm]{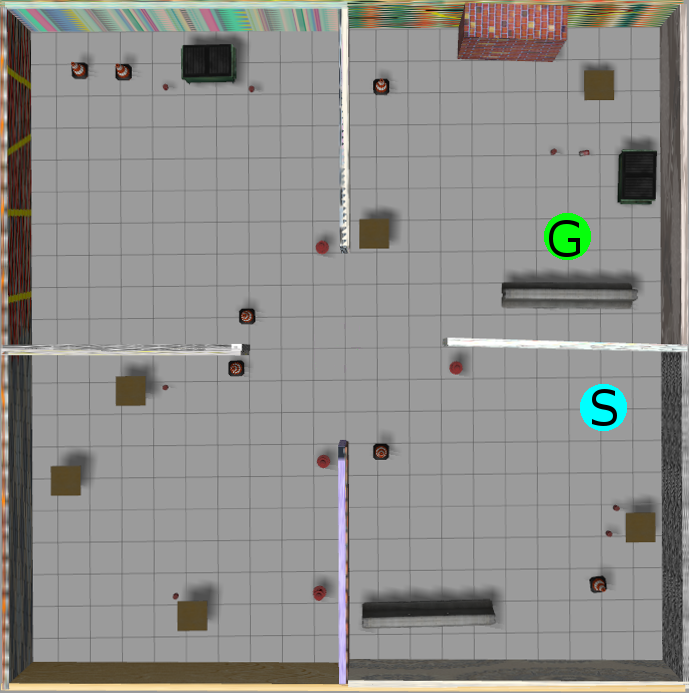}
	}
    \subfloat[]{
	\includegraphics[width=3.5cm, height=3.5cm]{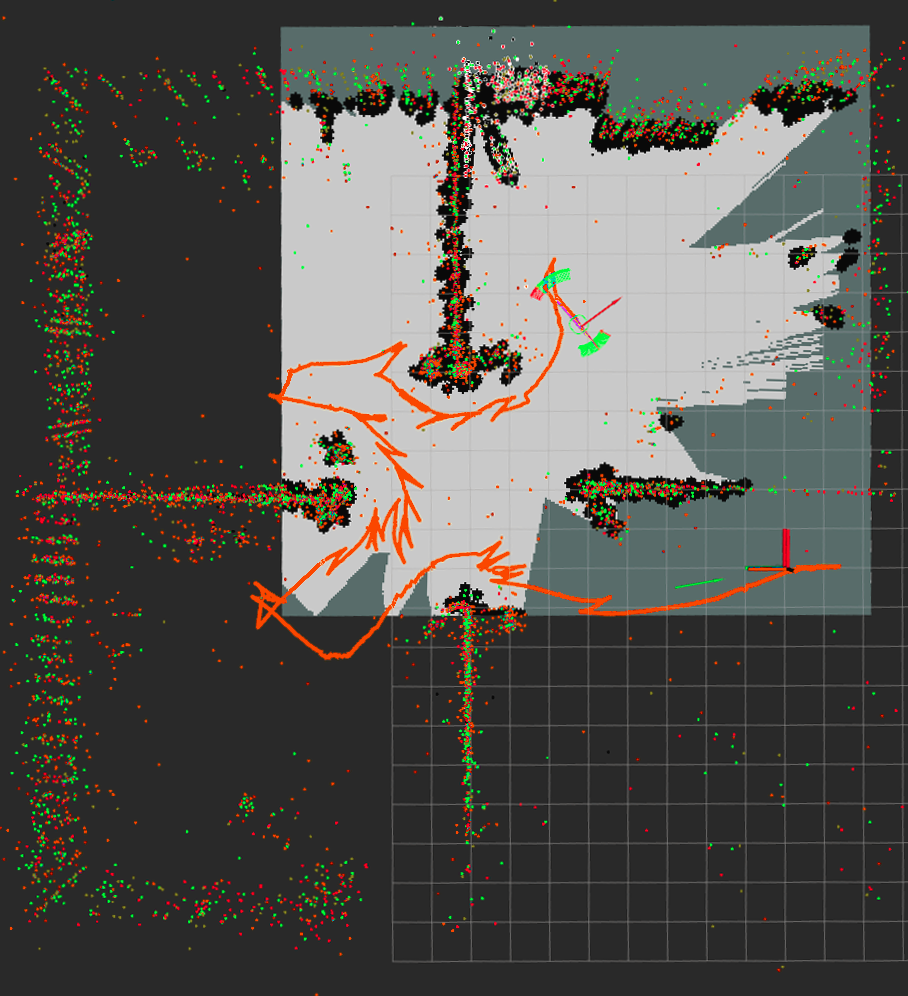} 
    }\\
	\caption{\small{Results obtained with ORB SLAM using the Navigation Stack on maps 5 (a, b) 6 (c, d) and 7(e, f). The first column shows the environments. The robot was made to navigate from the start location (S) to the goal location (G). The second column shows the reconstruction from ORB SLAM and the robot's estimated path after full exploration.}}
\label{orb_results}
\vspace{-3em}
\end{figure}

\begin{figure}
	\centering
	\subfloat[]{
	\includegraphics[width=3.5cm, height=3.5cm]{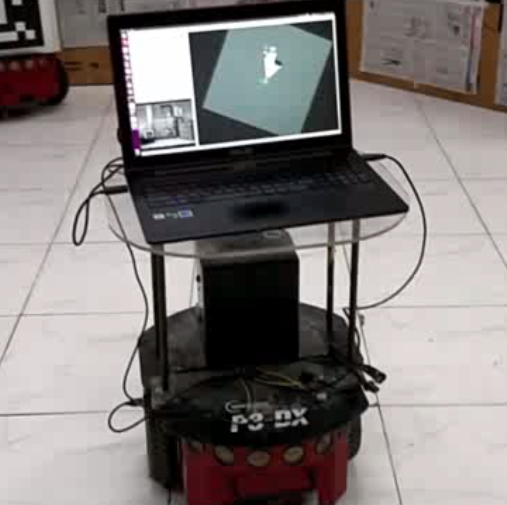}
	}
	\subfloat[]{
	\includegraphics[width=3.5cm, height=3.5cm]{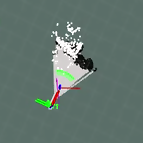}
	} \\
	
	\subfloat[]{
	\includegraphics[width=3.5cm, height=3.5cm]{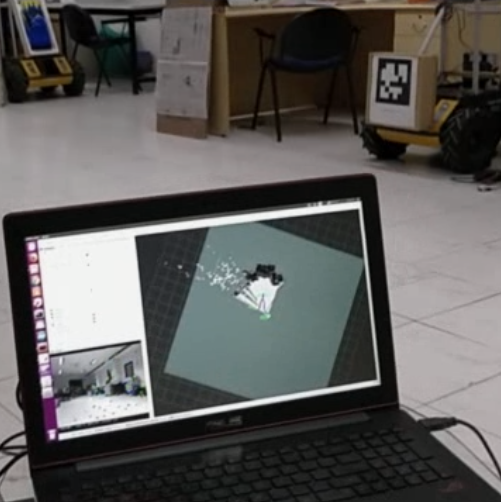}
	}
	\subfloat[]{
	\includegraphics[width=3.5cm, height=3.5cm]{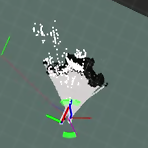} 
    }\\
    
    \subfloat[]{
	\includegraphics[width=3.5cm, height=3.5cm]{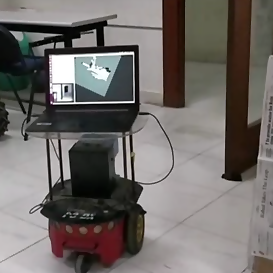} 
    }\subfloat[]{
	\includegraphics[width=3.5cm, height=3.5cm]{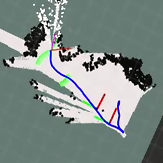}
	}\\
	
    \subfloat[]{
	\includegraphics[width=3.5cm, height=3.5cm]{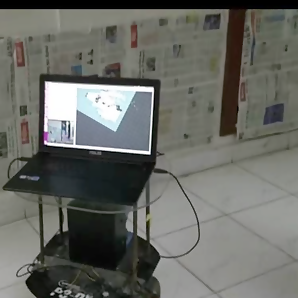}
	}
	\subfloat[]{
	\includegraphics[width=3.5cm, height=3.5cm]{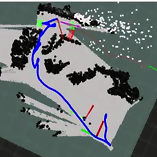} 
    }\\

	\caption{\small{Results of real time exploratory run using the RL based Planner. First row shows the robot during its navigation from a large room to a open corridor. The second row shows trajectory and map estimates using ORB SLAM with RL based Planner where the predicted failures followed by back and forth motion around the opening of the room can be seen (as shown in (g) and (h) respectively). }}
\label{Trial_Runs}
\vspace{-3.5em}
\end{figure}

In Fig. \ref{orb_results} we show the simulation environments and full maps recovered with ORB SLAM using the RL based Planner based on the ROS Navigation Stack. Until now, autonomous navigation with monocular camera has been done only for short distances. We specifically designed these simulation environments to be difficult for navigation, with larger map sizes(20mX20m) and sparse feature scenes like featureless walls. Goals were given in different parts of the map and the robot had to explore its way to the goal without any prior knowledge of the surrounding. Table \ref{results_table_nav_stack} shows the qualitative results obtained on Maps 5-7, where the RL based planner works considerably better than other methods. Exploration makes performing Monocular SLAM especially difficult due to the increased rotatory motion of the camera which could lead to distorted maps. Results of an exploratory run on a real robot where goals are user-specified are shown in Fig. \ref{Trial_Runs}. The robot is made to navigate from inside a room to the adjoining corridor. Such a transition in the environment is pretty challenging for simple Monocular SLAM.\par

\small
\begin{table}[H]
\vspace{-0.5em}
\caption{Number of Successful Runs (on Map 5-7) for Goal Based Navigation using Navigation Stack with ORB SLAM. These results can be seen in Fig. \ref{orb_results}}
\label{results_table_nav_stack}
\begin{center}
\vspace{-1em}
\begin{tabular}{|c|c|c|c|}
\hline
&&&\\
\textbf{Policy} & Map 5 & Map 6 & Map 7 \\
&(15 runs)&(15 runs)&(15 runs)\\
\hline
Naive & 3 & 2 & 2 \\
RL & \textbf{12} & \textbf{11} & \textbf{13}\\
\hline
Naive + Odom & 2 & 6 & 7\\
RL + Odom & \textbf{14} & \textbf{13} & \textbf{14} \\
\hline
Naive + Reloc + Odom & 6 & 11 & 11\\
RL + Reloc + Odom & \textbf{15} & \textbf{15} & \textbf{15}\\
\hline
\end{tabular}
\vspace{-1em}
\end{center}
\end{table}
\normalsize

To demonstrate the advantage of our approach over methods which use sensor fusion for localization, we augment SLAM's localization output with wheel odometry data obtained from Gazebo. Multi-modal sensor fusion is expected to alleviate some of the localization issues during navigation. However, as visible from the $2^{nd}$ and $3^{rd}$ row of Table \ref{results_table_nav_stack} (policies "RL" and "Naive + Odom" respectively), our RL based planner without wheel odometry performs better than a naive planner with sensor-fusion based localization output. Fusing our RL based Planner with the odometry is able to further improve the navigation capabilities as seen in the $4^{th}$ row of Table \ref{results_table_nav_stack} ("RL + Odom").

We also show the superiority of the RL based Planner against the ROS Navigation Stack which utilizes ORB SLAM's relocalization capabilities to restart SLAM after a tracking failure. When a failure occurs, the robot is manually given a goal facing a previously visible scene thus allowing relocalization. To allow the robot to keep moving after a tracking failure, the SLAM localization output is augmented with wheel odometry data. The $5^{th}$ and $6^{th}$ rows of Table \ref{results_table_nav_stack} ("Naive + Reloc + Odom" and "RL + Reloc + Odom"), show the results of ORB SLAM with relocalization. We allow only 2 relocalization attempts as increasing these attempts leads to degradation of the SLAM output while also increasing the time required to reach the goal. Finally, augmenting our RL based Planner with relocalization capabilities gives 100\% success, with accurate reconstruction as indicated in the last row of Table \ref{results_table_nav_stack}.


\section{\textbf{Conclusion and Future Work}}

We propose a map-agnostic Reinforcement Learning based navigational planner that learns to identify motions that can lead to Monocular SLAM failures and avoids them by performing recovery behaviors. The generated trajectories have minimal failures and pose estimation errors which is proved by carrying out experiments both in simulations and on real robots. This is different from traditional belief space planners or other SPLAM formulations due to their reliance on dense maps or voxel grids. In contrast, our pipeline works for sparse maps where the structure cannot be easily captured as compared to dense depth maps. \par 

Interestingly, the trajectories generated by the RL based planner show a significant number of back and forth motions near regions of low continuity of visible mapped points. This is similar to manual motions employed with Monocular SLAM while using handheld cameras. This allows the SLAM algorithm to have continuity in the features tracked before moving into a new environment, thus leading to a better SLAM estimate.\par 

The method is not only transferable to different maps but over different feature based Monocular SLAM algorithms as well. Values learned using one Monocular SLAM system are transferred to another, without any difficulties. In terms of learning SLAM failure safe behaviour, our methods showed about 37\% improvement in comparison with a state-of-the-art Next-best-view planner and about 45\% in comparison with Supervised Learning based planners. We also show the superiority of our approach over methods which use use wheel odometry for multi-sensor fusion or relocalization. \par

In the future, we look forward to improving the performance of the system by integrating more recent techniques in Deep Reinforcement Learning and Deep Policy Gradients, which would allow us to add more parameters in the equation and explore into continuous space planning. Further work can also be carried out in studying how well these method scales to direct Monocular SLAM methods. Moreover, this method can be scaled to aerial robots and learn SLAM failure safe behaviour for drones. In such cases, one could explicitly make use of the fact that better baselines between views can be obtained due to the horizontal/vertical motion capabilities of a drone as compared to just the forward/backward motion in case of ground robots. \par

\bibliographystyle{ACM-Reference-Format}
\bibliography{references}

\end{document}